\pdfoutput=1

\documentclass[11pt]{article}

\usepackage[x11names]{xcolor}
\usepackage{acl}
\usepackage[shortlabels]{enumitem}
\usepackage{times}
\usepackage{color,soul} 
\usepackage{booktabs}
\usepackage{enumitem}

\usepackage{latexsym}
\usepackage{hyperref}
\usepackage{multirow}
\usepackage[LAE,T1]{fontenc}
\usepackage{arabtex}
\usepackage{utf8} 
\usepackage[utf8]{inputenc} 
\usepackage[LAE,T1]{fontenc}
\usepackage[arabic,french,english]{babel}
\usepackage[raggedrightboxes]{ragged2e}
\usepackage{todonotes}
\usepackage{array}
\usepackage{lscape}
\usepackage{natbib}
\usepackage{longtable}
\usepackage{multirow}
\usepackage{float}
\usepackage{makecell}
\usepackage{xcolor, colortbl}
\usepackage{inconsolata}
\usepackage{arydshln}
\usepackage{array}
\usepackage{soul}
\usepackage{times}
\usepackage{latexsym}

\usepackage{multirow}

\usepackage{microtype}
\usepackage{hyperref}

\usepackage{microtype}
\usepackage{rotating}
\usepackage{amsmath}

\usepackage{inconsolata}

\usepackage{graphicx}
\usepackage{subcaption}
\usepackage{multirow}
\usepackage{tabularx}
\usepackage{framed}

\setcode{utf8}

\title{To Distill or Not to Distill? \\ On the Robustness of Robust Knowledge Distillation}

\author{
  Abdul Waheed$^{\xi}$\,
  ~~~~~Karima Kadaoui$^{\xi}$\,
  ~~~~~Muhammad Abdul-Mageed$^{\xi,\gamma,\lambda}$ \\
  $^{\xi}$MBZUAI~~~~~ 
  $^{\gamma}$The University of British Columbia ~~~~~$^\lambda$ Invertible AI \\
  \texttt{\normalsize \{abdul.waheed,karima.kadaoui\}@mbzuai.ac.ae}~~~~~~~~\texttt{\normalsize muhammad.mageed@ubc.ca}
}

\begin{document}
\maketitle
\section*{~~~~~~~~~~~~~~~~~~~~~~~~~~~~~Abstract}
Arabic is known to present unique challenges for Automatic Speech Recognition (ASR). On one hand, its rich linguistic diversity and wide range of dialects complicate the development of robust, inclusive models. On the other, current multilingual ASR models are compute-intensive and lack proper comprehensive evaluations. In light of these challenges, we distill knowledge from large teacher models into smaller student variants that are more efficient. We also introduce a novel human-annotated dataset covering five under-represented Arabic dialects for evaluation. We further evaluate both our models and existing SoTA multilingual models on both standard available benchmarks and our new dialectal data. Our best-distilled model's overall performance ($45.0$\% WER) surpasses that of a SoTA model twice its size (SeamlessM4T-large-v2, WER=$47.0$\%) and its teacher model (Whisper-large-v2, WER=$55.1$\%), and its average performance on our new dialectal data ($56.9$\% WER) outperforms all other models. To gain more insight into the poor performance of these models on dialectal data, we conduct an error analysis and report the main types of errors the different models tend to make. The GitHub repository for the project is available at \url{https://github.com/UBC-NLP/distill-whisper-ar}.

\section{Introduction}\label{introduction} 

There have been significant advancements in multilingual automatic speech recognition (ASR) in both training methodologies and architectures. Models such as OpenAI's Whisper~\cite{whisper} and Meta's SeamlessM4T~\cite{Communication2023SeamlessM4TMM} can transcribe speech from languages in the order of the hundreds, albeit with varying degrees of accuracy. Especially for low-resource languages, these models do not perform well~\cite{whisper, williams23_sigul, talafha2023n}. Arabic, for example, poses significant challenges to these multilingual models and hence is the object of the current work.

Arabic can be classified into three broad categories, namely: \textbf{Classical Arabic} (CA), used in early literature and religious texts; \textbf{Modern Standard Arabic} (MSA), the `high' variety used in official documents and in the media; and \textbf{Dialectal Arabic} (DA), the collection of `low' varieties used in day-to-day conversations~\cite{varieties}. DA can vary extensively at the regional level (e.g. Gulf vs Maghrebi), country level (e.g. Egyptian vs Sudanese), and sub-country (e.g. Hourani or Northern Jordanian Dialect vs Urban or Madani dialect)~\cite{bookhabash, abdul-mageed-etal-2020-toward, Shon2020ADI17AF, abdul-mageed-etal-2018-tweet}. Due to the significant differences in lexicon, phonetics, and even grammar between these varieties, ASR systems trained on MSA alone cannot be reliably leveraged off-the-shelf for all Arabic speech.
Developing effective models for DA can prove especially difficult, given the lack of standardized orthography, the scarceness of labeled data for many dialects, inconsistent use of diacritics, and use of code-switching~\cite{ali2021arabic}.

Although most multilingual and multimodal systems (e.g., ~\cite{whisper, m4t, Communication2023SeamlessM4TMM}) cover Arabic, their evaluation predominantly involves benchmarks established for MSA, such as FLEURS \cite{conneau2022fleurs}, Common Voice (CV) \cite{ardila2020common}, and the Arabic Speech Corpus (ASC)~\cite{asc}. 
Since Arabic exhibits substantial linguistic diversity, encompassing various varieties and dialects, evaluations conducted solely on MSA are inherently limited. 
Existing works aiming to address this gap, e.g.,~\cite{talafha2023nshot}, lack thorough evaluation and do not cover current state-of-the-art (SoTA) models. To address this, we conduct a comprehensive evaluation of all recently developed models on a linguistically diverse set of Arabic datasets. 

Beyond the challenge of inadequate evaluation, the deployment of massive multilingual multimodal systems such as SeamlessM4T~\cite{Communication2023SeamlessM4TMM} and Whisper~\cite{whisper} is hampered by the considerable computational resources they require during both training and inference. These efficiency issues pose a significant accessibility barrier, discriminating against populations with limited resources. To alleviate this concern, we employ a framework for knowledge distillation~\cite{gandhi2023distilwhisper} from large models such as Whisper~\cite{whisper} into relatively compact models for Arabic speech recognition. We show that our distilled models are not only compute-efficient but their performance is on par or better compared to larger counterparts.

In summary, the gaps in existing work include (1) the insufficient knowledge about the utility of recent multilingual speech model models on Arabic, including dialects, (2) the discrepancy in representing some Arabic dialects in existing dialectal benchmarks, and (3) the inefficiency of these models due to their large sizes which demands significant compute resources at both training and inference time. We address these limitations through a number of contributions, as follows: 

\begin{itemize} 
    \item We evaluate major multilingual speech models on a wide variety of standard benchmarks representing Arabic to identify their zero-shot performance.
    \item To evaluate the models under diverse varieties, we introduce a never-seen in-house labeled ASR dataset covering five under-represented Arabic dialects.
    \item We distill knowledge from large ASR models into relatively small, and hence more efficient, (student) models with minimal-to-no performance drops compared to the bigger (teacher) counterparts.
\end{itemize}

The rest of the paper is organized as follows: Section~\ref{related-work} is a review of related works.
In Section~\ref{knowledge-distillation}, we introduce knowledge distillation and outline our related methods and training strategies. In Section~\ref{experiments}, we provide details about our experiments, and in Section~\ref{results} introduce and discuss our results. Section~\ref{error-analysis} delivers a thorough error analysis based on the model predictions on our new dialectal data. We conclude the work in Section~\ref{conclusion}. Finally, we oultine our limitations and ethical considerations in Sections~\ref{limitations} and~\ref{ethics-statement}, respectively.
\section{Related Work}\label{related-work}

\textbf{Multilingual ASR.} Recent efforts in ASR have focused on building massive multilingual systems~\cite{Communication2023SeamlessM4TMM, m4t, whisper, MMS, dhawan2023towards, Rekesh2023FastCW, zhang2023google, baevski2020wav2vec, conneau2020unsupervised}. These multilingual models perform quite well for high-resource languages such as English across various evaluation settings. However, they often perform poorly for low-resource languages and in challenging settings~\cite{williams23_sigul, talafha2023n, Chemudupati2023OnTT, bhogale2023vistaar, article, whisper}. This suggests that a thorough evaluation of these systems for low-resource languages is needed. 

\textbf{Arabic ASR.} For Arabic, the performance of these models remains under-explored. While OpenAI's Whisper model~\textit{whisper-large-v3}~\cite{whisper} achieves 15.1\% word error rate (WER) on Common Voice 15.0's~\cite{ardila2020common} Arabic split and 9.6\% WER on FLEURS~\cite{conneau2022fleurs}, a performance close to human-level,~\citet{talafha2023nshot} show that it is vulnerable to linguistic variations where its performance degrades substantially on several Arabic dialects. Furthermore, the performance of other multilingual systems such as SeamlessM4T~\cite{Communication2023SeamlessM4TMM}, Universal Speech Model (USM)~\cite{zhang2023google}, and XLS-R~\cite{xls-r} on diverse Arabic varieties remains unknown.

\textbf{Efficiency.} The size of these massive multilingual systems poses another challenge to their usability. To address this, \citet{speculative_decoding_for_2x_faster_whisper_inference} shows that speculative decoding~\cite{leviathan2023fast} can expedite the generation from Whisper by a factor of two. Efficient transformer inference engines such as CTranslate2~\cite{CTranslate2} based inference~\citet{faster_whisper} can also improve the generation speed, despite having the same memory requirements. Moreover, although quantization techniques have been effective in reducing memory requirements~\cite{frantar2022gptq, lin2023awq}, they do not decrease the number of active parameters, leading to variable improvement in generation speed~\cite{jin2024comprehensive}.

\textbf{Knowledge distillation.} Knowledge distillation is a method used to transfer knowledge from large models to smaller ones, thereby reducing both memory and compute requirements~\cite{hinton2015distilling, sanh2019distilbert, Gou_2021, lopes2017datafree, kim-rush-2016-sequence}. This technique has been effectively applied in various domains. For example, in computer vision applications, knowledge distillation results in compact and efficient models~\cite{kaleem2024comprehensive, compress}. Similarly, in diffusion models~\cite{Luo2023ACS} and large language models~\cite{xu2024survey}, knowledge distillation produces small, efficient, and task-specific models. ~\citet{yang2023knowledge} distill knowledge from multiple foundation models into small and dedicated speech recognition models. ~\citet{ni2023adaptive} proposes cross-modality knowledge distillation from large language models into speech models.

\textbf{Knowledge distillation in speech.}~\citet{ferraz2024multilingual} distill knowledge from a large Whisper model into small multilingual models but limit their evaluation to standard benchmarks in eight languages (Arabic not included in the set).~\citet{shao2023whisperkdq} apply a novel distillation approach to Whisper, reducing its size by 80-90\% while also improving its performance. \citet{distilhubert} propose a layer-wise distillation approach that reduces the size of a Hubert model by 75\% while increasing its processing speed by 73\%, retaining most of the original model’s performance across multiple tasks. In addition to that, researchers have introduced methods for model compression, such as data-free knowledge distillation and teacher-student (TS) learning for domain adaptation~\cite{data-free-kd, ts-learning}. These approaches involve training student models to mimic teacher models using various strategies, including Gaussian noise generation and sequence-level Kullback-Leibler (KL) divergence~\cite{kl-div}.

Among different knowledge distillation approaches such as the ones highlighted above, the standard student-teacher distillation is a task- and modality-independent framework that is simple yet effective.~\citet{gandhi2023distilwhisper} use this framework to distill Whisper into small monolingual models for English using large-scale pseudo-labels. However, their work is limited to high-resource language. We take inspiration from~\cite{gandhi2023distilwhisper} and distill Whisper into small models for Arabic and perform a thorough evaluation. One difference between our work and that of~\cite{gandhi2023distilwhisper} is that while there is limited information about the out-of-distribution datasets of ~\cite{gandhi2023distilwhisper}'s work and whether they are part of the teacher's training data, we employ new dialectal speech data never seen by the model. 

\section{Knowledge Distillation}\label{knowledge-distillation}
Knowledge distillation is a method of transferring knowledge from a large model (teacher) to a relatively small model (student). The student model is trained to mimic the behavior of the teacher model both at the dense representation level and the sequence level~\cite{hinton2015distilling, sanh2020distilbert, kim-rush-2016-sequence}. Following~\citet{gandhi2023distilwhisper}, who distill a Whisper model for English ASR, we first generate large-scale pseudo-labels from the teacher model and apply a threshold to filter the output. We then train the student model with high-quality filtered pseudo-labels as ground truth, which can be expressed as:
\begin{align}\label{eq1}
    \mathcal{L}_{PL} = - \sum_{i=1}^{N'} P\left( y_{i} | 
    \hat{\boldsymbol{y}}_{<i}, \boldsymbol{H}_{1:M} \right)
\end{align}

\noindent The student model is also trained to minimize the discrepancy between the probability distributions over tokens of the student and teacher models, based on KL divergence:
\begin{align}\label{eq2}
    \mathcal{L}_{KL} &= \sum_{i=1}^{N} KL \left(Q_{i}, P_{i} \right) 
\end{align}
\textbf{Objective}: We take the weighted sum of (\ref{eq1}) and (\ref{eq2}) to get the final objective, which can be written as:
\begin{align}\label{}
    \mathcal{L}_{KD} = \alpha_{KL} \mathcal{L}_{KL} + \alpha_{PL} \mathcal{L}_{PL}
\end{align}
We use the same values for $\alpha_{KL}$ ($0.8$) and $\alpha_{PL}$ ($1.0$) as~\citet{gandhi2023distilwhisper}. Details about our teacher and student models, along with training data, can be found in Table~\ref{distilled-models}.

\section{Experiments}\label{experiments}

\subsection{Datasets}\label{datasets}
\noindent\textbf{Common Voice.} CV~\cite{ardila2020common} is a widely used multilingual benchmark for speech recognition. In our experiments, we use the \textit{test} and \textit{validation} splits of four different CV versions (6.1, 9.0, 11.0, 15.0) which have been widely used in other work for training and evaluating Arabic ASR models~\cite{talafha2023nshot, waheed-etal-2023-voxarabica}. Upon inspection of the data, we found it to be composed of mostly MSA along with some CA speech. 

\noindent\textbf{Multi-Genre Broadcast.} Multi-genre broadcast (MGB)~\cite{ali2019mgb2, 9003960, ali2017speech} is a challenge for a wide range of Arabic speech understanding tasks such as speech recognition, speaker identification, dialect identification, etc. We experiment with three variants, namely MGB2, MGB3, and MGB5. MGB2 has roughly around $70\%$ MSA, with the remainder containing other dialects~\cite{mgb2}. MGB3 is predominantly composed of Egyptian Arabic, while MGB5 focuses on Moroccan Arabic.

\noindent\textbf{FLEURS.} FLEURS~\cite{conneau2022fleurs} is a multilingual collection of parallel speech corpora. We use the \textit{dev} and \textit{test} splits of the Arabic subset ``\texttt{ar\_eg}'', which contains MSA spoken with an Egyptian accent, to evaluate our models in a zero-shot setting. We also use the \textit{train} split in distillation.

\noindent\textbf{In-House Data.} 
In response to the notable scarcity of publicly available dialectal data, we manually curate a dataset representing five underrepresented Arabic dialects, namely Algerian (ALG), Jordanian (JOR), Palestinian (PAL), Emirati (UAE), and Yemeni (YEM), spanning four dialectal regions (North African, Levantine, Gulf, and Yemeni).
We task native speakers of each dialect to annotate segments from local TV series sourced from YouTube. Our dataset comprises a total of $10,567$ utterances and 	$121,293$ words ($2,133$ utterances and $24,258$ words per dialect, on average) amounting to over $13$ total hours. Individual statistics for each dialect can be found in Table \ref{tab:home-data-stats}.

\begin{table}[]
    \centering
    \renewcommand{\arraystretch}{1.}   
    \resizebox{1.\linewidth}{!}{

    \begin{tabular}{lcccc}
        \toprule
        Dia. & \makecell{Utt.} & \makecell{Words} & \makecell{Words/Utt.} & \makecell{Hours} \\
        \midrule
        ALG & 815 & 8,900 & 10.92 & 0.97 \\
        JOR &	2,671 & 28,291 & 10.59 & 3.27\\
        PAL & 1,097 & 15,152 & 13.81 & 1.67 \\
        UAE & 3,701 & 41,345 & 11.17 & 4.42 \\
        YEM & 2,283 & 27,605 & 12.09 & 2.94 \\ 
        \hdashline
        \textbf{Total} & 10567 & 121293 & 11.48 & 13.29 \\ \hdashline
        \textbf{Avg.} & 2113.4 & 24258.6 & 11.72 & 2.65 \\
        \bottomrule
    \end{tabular}
    }
    \caption{Utterance and word count statistics across the different dialects from our in-house dataset.\hspace{\textwidth}Di.: Dialect. \#: Number of. \textbf{Avg}.: Average. \textbf{Utt}.: Utterance}
    \label{tab:home-data-stats}
\end{table}

\subsection{Models}
We evaluate a wide range of multilingual speech recognition models on different varieties of Arabic from the aforementioned datasets, including standard and accented MSA, and various Arabic dialects. We also distill small dedicated\footnote{Our models are `dedicated' in the sense that they are solely focused on Arabic and only handle ASR.} models from larger Whisper models. We categorize these systems as follows:

\subsubsection{Supervised Baselines}
  We evaluate two openly available supervised baselines along with a Whisper model that we fine-tune on Arabic ASR in a supervised setting. The first two models are \href{https://huggingface.co/AndrewMcDowell/wav2vec2-xls-r-1b-arabic}{Wav2Vec2-XLS-R}~\cite{conneau2020unsupervised, babu2021xlsr}, trained on CV8.0 which has significant overlap with other versions of the CV dataset, and \href{https://huggingface.co/omarxadel/hubert-large-arabic-egyptian}{HuBERT}~\cite{hsu2021hubert}, trained on MGB-3~\cite{ali2017speech} and the Egyptian Arabic Conversational Speech Corpus ($5.5$ hours). The third model is \textit{whisper-large-v2}, which we fine-tune on CV11.0 and MGB-2. We evaluate all three models on the datasets listed in Section~\ref{datasets}. This includes the in-distribution \textit{test} and \textit{dev} splits of MGB-2 and CV11.0. 

\subsubsection{Zero-Shot Models}
Large multilingual speech models are acclaimed for transcending language and task barriers. In particular, these models are usually claimed to demonstrate proficiency in a variety of speech tasks on English in the \textit{zero-shot} setting. However, it is crucial to conduct thorough evaluations of these models on other languages and dialects and under diverse conditions. Hence, our objective is to assess a wide array of zero-shot models on a wide range of Arabic speech recognition datasets to asses their robustness and generalization capability beyond English. We focus on a number of recently introduced models that have gained popularity in the community as well as existing commercial systems, as wel explain next.

\noindent\textbf{Whisper.} Whisper~\cite{whisper} is a multilingual speech model capable of speech recognition and translation across languages including Arabic. We evaluate four variants of Whisper, namely \href{https://huggingface.co/openai/whisper-small}{\textit{small} (W-S)}, \href{https://huggingface.co/openai/whisper-medium}{\textit{medium} (W-M)}, \href{https://huggingface.co/openai/whisper-large-v2}{\textit{large-v2} (W-L-v2)}, and \href{https://huggingface.co/openai/whisper-large-v3}{\textit{large-v3} (W-L-v3)}. We use all the default parameters for decoding with a maximum sequence length of 225 tokens. 

\noindent\textbf{SeamlessM4T.} 
Multimodal multilingual speech models are also capable of generating high-quality transcripts across languages~\cite{Communication2023SeamlessM4TMM}. However, they lack a comprehensive evaluation in languages besides English. We address this by evaluating three available variants of SeamlessM4T (\href{https://huggingface.co/facebook/seamless-m4t-medium}{\textit{medium} (SM4T-M)}, \href{https://huggingface.co/facebook/seamless-m4t-large}{\textit{large-v1}(SM4T-L-v1)} and \href{https://huggingface.co/facebook/seamless-m4t-v2-large}{\textit{large-v2} (SM4T-v2)}) for Arabic ASR in a zero-shot setting. We use all the default parameters provided in the model's inference pipeline.


\noindent\textbf{Commercial Systems.}
We broaden our evaluation beyond publicly accessible ASR models, incorporating proprietary platforms, with a focus on Amazon's ASR system. Due to cost considerations, our evaluation is exclusively centered on the Amazon Transcribe service on our in-house data.\footnote{\url{https://aws.amazon.com/transcribe/}}

\subsubsection{Distilled Models}\label{distilled-models}
As described in Section \ref{knowledge-distillation}, we distill \textit{whisper-large-v2} into seven different student models (see Table~\ref{distilled-models}). We provide more details about the teacher and student models and distillation data here. 

\noindent\textbf{Teacher and Student Models.} We use a \href{https://huggingface.co/openai/whisper-large-v2}{\textit{whisper-large-v2} checkpoint} for pseudo-labeling and the same model as the teacher during training. We train four variants of the student model in different configurations in terms of the number of layers being removed. Following~\newcite{gandhi2023distilwhisper}, we initialize the student models with maximally spaced layers in the encoder and decoder block of the teacher model. We provide more details about our distilled models in Table~\ref{distilled-models}.


\begin{table}[h]
\centering
\begin{tabular}{lcccc}
\toprule
\multicolumn{1}{l}{\textbf{Model}} & \multicolumn{1}{l}{\# EL} & \multicolumn{1}{l}{\# DL}  & \multicolumn{1}{l}{Data} \\ \midrule
W-L-v2                           & 32                                                                                    & 32 & N/A                                                                                \\ \hdashline
DW-8-8                         & 8  & 8 & 100K  \\
DW-16-16                       & 16 & 16 & 100K \\
DW-32-16                      & 32 & 16  & 100K \\
DW-16-32                     & 16 & 32 & 100K  \\ \bottomrule 
DW-16-16++                   & 16 & 16 & 500K \\
DW-32-16++                     & 32 & 16 & 500K \\ \hdashline
DW-16-16-1M                     & 32 & 16 & 1M \\ \bottomrule
\end{tabular}
\caption{ \label{distilled-models}
The student models are initialized from maximally spaced layers of the teacher model. The size of data is stated as the number of segments. All distilled models are trained for ten epochs. \textbf{W-L-v2:} Whisper-large-v2. \textbf{\#:} Number of. \textbf{DW:} Distill-Whisper. \textbf{EL:} Encoder Layers. \textbf{DL:} Decoder Layers.}

\end{table}

\noindent\textbf{Training Data.} We randomly sample 100K and 500K segments from a mixture of MGB2~\cite{mgb2}, MGB3~\cite{ali2017speech}, FLEURS~\cite{conneau2022fleurs}, CommonVoice 15.0~\cite{ardila2020common}, QASR~\cite{mubarak2021qasr}, Arabic Speech Corpus~\cite{asc}, and Massive Arabic Speech Corpus (MASC)~\cite{e1qb-jv46-21}. This amounts to roughly 100 and 500 hours of pseudo labeled speech data, respectively. We explicitly include only the train split of each dataset.  

\subsection{Experimental Setup}
We conduct all of our training and evaluation experiments on 8xA100/4xA100 (40G) GPU nodes. For the evaluation, we use the default decoding parameters used in the corresponding models unless otherwise specified. We use $225$ as the maximum sequence length throughout our experiments and report Word Error Rate (WER) and Character Error Rate (CER) as our evaluation metrics. 
For distillation, we use a value of 80\% for the WER threshold $\lambda$ to filter-out low-quality transcription from pseudo-labels for the results reported in Table~\ref{main-results}. We also experiment with different threshold values and discuss the findings in Section~\ref{subsec:threshold}. Although our threshold for main results seems too high, \citet{gandhi2023distilwhisper} find that going from a threshold of 80 to five yields a marginal improvement of one point in terms of average WER across different in-distribution and out-of-distribution evaluation sets. In addition, we believe that a high threshold value also helps approximate the performance where we do not have labeled data to conduct the filtering process, especially when labeled data is scarce. Due to computing limitations, we do not conduct any training hyperparameter search and directly apply the configuration used in~\newcite{gandhi2023distilwhisper}. For the distillation process, we report our key parameters in Table~\ref{training-parameters} (Appendix~\ref{training}).

\noindent\textbf{Text Preprocessing.}\label{text-preprocessing} 
In everyday writing, Arabic is characterized by inconsistencies in diacritics use and letter variations (e.g. {\small\<أ>} vs {\small\< ا>} ). This linguistic variability poses a challenge for ASR evaluation, as transcriptions that are phonetically accurate and intelligible to a native speaker might still be marked as errors due to strict lexical mismatches. To address this, we follow~\citet{talafha2023nshot, chowdhury2021model} to standardize and normalize the text. Specifically, we (1) remove any special characters and diacritics, (2) remove all Latin characters since we are not concerned about code-switching, (3) transliterate all Arabic digits (i.e. \<١>, \<٢>, \<٣>) to Arabic numerals (i.e. 1, 2, 3), and (4) normalize all \textit{alef} variations to the one with no \textit{hamza}. 
\section{Results and Discussion}\label{results}

\begin{table*}[h!]
\centering
\Large 
\renewcommand{\arraystretch}{1.1}   
\resizebox{1.\linewidth}{!}{
\begin{tabular}{clcccccccccccccc}
\toprule
& \multirow{2}{*}{Model} & \multirow{2}{*}{Size}  & \multirow{2}{*}{CV15.0} & \multirow{2}{*}{MGB2} & \multirow{2}{*}{MGB3} & \multirow{2}{*}{MGB5} & \multirow{2}{*}{Fleurs} & \multicolumn{5}{c}{In-house Data} & \multirow{2}{*}{Avg.}                                     \\
                       &               &          &                       &                       &                       &                         & & ALG          & JOR         & PAL         & UAE          & YEM          \\ \midrule
\multirow{16}{*}{\rotatebox{90}{Normalized + No Diacritics}} & Amazon & -/- & -/- & -/- & -/- & -/- & -/- & 83.6/70.2 & 45.5/25.6 & 52.4/29.0 & 58.8/40.8 & 64.7/43.5 & 61.0/41.8 \\ \cdashline{2-16}

& XLS-R & 0.96 & 89.7/39.4 & 97.6/53.1 & 98.7/61.6 & 99.5/68.0 & 94.9/43.9 & 99.7/67.0 & 99.1/61.4 & 99.1/61.1 & 99.4/64.6 & 99.5/63.6 & 97.7/58.4 \\

& HuBERT & 0.31 & 55.2/18.9 & 49.6/17.3 & \textbf{25.2/9.5} & 92.4/45.5 & 34.9/10.9 & 96.8/44.3 & 65.2/23.3 & 73.8/27.9 & 83.0/36.7 & 90.5/38.8 & 66.7/27.3 \\

& W-FT & 1.5 & 35.8/21.9 & \textbf{15.3/8.1} & 48.9/26.9 & 101.4/62.3 & 9.8/3.4 & 115.5/69.6 & 67.8/37.2 & 69.6/35.4 & 105.9/69.1 & 107.1/64.8 & 67.7/39.9\\ \cline{2-16}

& MMS-all & 1.0 & 106.4/80.9 & 39.3/13.4 & 75.3/34.6 & 89.7/45.9 & 23.8/6.3 & 100.2/78.0 & 89.8/55.4 & 99.9/75.1 & 100.1/78.1 & 100.2/76.6 & 82.5/54.4 \\ \cdashline{2-16}

& SM4T-M & 1.2 & 16.3/5.7 & 19.5/9.0 & 41.4/21.7 & 83.8/46.6 & 8.7/3.6 & 81.1/39.7 & 46.3/15.9 & 55.2/20.1 & 59.8/24.7 & 68.9/29.5 & 48.1/21.7 \\

& SM4T-L-v1 & 2.3 & 19.8/7.3 & 21.8/10.5 & 44.4/22.6 & 89.9/52.1 &  11.1/5.1 & 87.9/47.8 & 50.7/18.8 & 57.5/23.1 &  61.8/27.4 & 72.2/32.5 & 51.7/24.7 \\

& SM4T-L-v2 & 2.3 & \textbf{11.3/3.5} & 17.3/8.7 & 36.2/18.6 & 89.1/53.7 & \textbf{7.6}/4.0 & 92.1/52.0 & \textbf{41.5}/14.6 & \textbf{49.5/17.2} & 55.9/23.3 & 69.7/30.7 & 47.0/22.6 \\ \cdashline{2-16}

& W-S & 0.24 & 40.3/16.4 & 46.8/24.7 & 81.4/51.9 & 226.5/164.8 & 28.2/8.7 & 130.7/84.7 & 68.6/32.9 & 73.8/36.3 & 97.8/59.7 & 107.1/66.7 & 80.8/45.7 \\

& W-M & 0.77 & 29.8/13.2 & 33.1/18.5 & 64.3/39.5 & 127.7/88.3 & 16.4/5.1 & 103.7/69.9 & 50.5/21.1 & 58.7/24.7 & 82.5/52.6 & 86.8/52.0 &  65.4/38.5\\

& W-L-v2 & 1.5 & 19.6/7.8 & 26.5/15.3 & 53.0/33.0 & 99.2/68.9 & 11.4/3.6 & 106.4/71.7 & 42.3/17.0 & 51.1/22.3 & 63.8/38.2 & 77.3/45.5 & 55.1/32.3 \\

& W-L-v3 & 1.5 & 15.8/5.2 & 15.9/7.6 & 35.7/17.3 & 79.8/44.6 & 9.7/3.2 & 101.9/65.4 & 43.6/16.3 & 53.4/22.7 & 63.4/32.7 & 76.1/38.9 & 49.5/25.4 \\ \cline{2-16}

& DW-8-8 & 0.44 & 32.7/12.3 & 39.6/17.8 & 64.9/36.6 & 89.7/53.0 & 29.8/11.4 & 91.4/48.2 & 66.2/29.0 & 73.2/33.0 & 78.0/38.4 & 82.9/41.5 & 64.8/32.1 \\

& DW-16-16 & 0.80 & 22.1/7.2 & 26.0/10.8 & 50.5/25.1 & 82.4/43.3 & 18.8/6.6 & 83.0/38.5 & 50.4/18.2 & 61.0/23.3 & 64.6/27.7 & 72.7/31.6 & 53.2/23.2 \\

& DW-32-16 & 1.12 & 18.8/5.9 & 21.1/8.9 & 43.8/21.4 & 78.9/\textbf{40.4} & 14.2/4.8 & 79.5/33.4 & 44.4/14.7 & 55.0/19.5 & 58.1/22.8 & 68.5/28.1 & 48.2/20.0 \\

& DW-16-32 & 1.22 & 21.5/7.3 & 25.0/10.7 & 49.1/26.3 & 83.0/47.5 & 18.4/6.0 & 84.3/44.0 & 49.8/18.0 & 60.3/25.4 & 64.4/29.0 & 73.8/36.8 & 53.0/25.1\\ 

\cdashline{2-16}
& DW-16-16++ & 0.80 & 19.2/6.2 & 23.0/10.2 & 47.2/24.8 & 79.0/42.6 & 15.0/5.2 & 79.0/39.0 & 46.7/17.2 & 56.4/21.6 & 60.4/26.8 & 69.1/31.5 & 49.5/22.5 \\
& DW-32-16++ & 1.12 & 17.1/5.5 & 19.7/8.8 & 40.7/20.3 & \textbf{76.6}/40.6 & 11.1/\textbf{3.1} & \textbf{74.6/33.3} & 41.6/\textbf{13.4} & 51.4/18.8 & \textbf{53.5/21.1} & \textbf{63.5/26.8} & \textbf{45.0/19.2} \\

\hline
\multirow{16}{*}{\rotatebox{90}{Orthographic}} & Amazon & -/- & -/- & -/- & -/- & -/- & -/- & 88.0/71.6 & 59.2/29.1 & 63.4/32.2 & 71.1/44.3 & 77.4/47.7 & 71.8/45.0 \\ 
\cdashline{2-16}

& XLS-R & 0.96 & 92.7/46.7 & 97.7/54.5 & 99.1/64.5 & 99.6/70.1 & 95.1/45.4 & 99.7/68.0 & 99.3/62.9 & 99.2/62.8 & 99.5/66.4 & 99.7/66.4 & 98.2/60.8 \\

& HuBERT &0.31 & 76.5/31.0 & 59.4/20.3 & \textbf{43.3/16.5} & 95.0/48.7 & 48.9/14.4 & 96.2/45.6 & 70.6/25.4 & 81.5/31.4 & 87.9/39.9 & 91.3/40.8 & 75.1/31.4 \\

& W-FT & 1.5 & 70.0/33.8 & 29.4/10.9 & 60.1/32.2 & 105.0/64.3 & 28.7/7.3 & 114.5/70.3 & 75.1/39.0 & 81.3/38.7 & 113.7/70.9 & 110.1/65.6 & 78.8/43.3 \\ \cdashline{2-16}

& MMS-all & 1.0 & 106.0/82.5 & 40.3/14.0 & 77.7/38.1 & 90.4/48.5 & 28.8/7.8 & 100.2/77.8 & 91.5/56.2 & 100.0/75.8 & 100.1/78.4 & 100.1/76.8 & 83.5/55.6 \\ \cdashline{2-16}

& SM4T-M & 1.2 & 42.3/18.2 & 28.1/11.2 & 50.2/26.8 & 88.2/50.8 & 19.5/6.0 & 84.5/42.8 & 55.2/18.7 & 63.0/23.0 & 68.0/28.1 & 79.4/34.5 & 57.8/26.0 \\

& SM4T-L-v1 & 2.3 & 44.2/19.1 & 25.9/11.7 & 52.5/27.6 & 92.8/55.9 & 22.6/7.6 & 89.7/50.3 & 59.1/21.7 & 64.7/25.8 & 69.0/30.3 & 81.5/37.0 & 60.2/28.7 \\

& SM4T-L-v2 & 2.3 & \textbf{37.7/15.8} & 22.4/9.9 & 46.7/23.9 & 92.1/58.4 & 19.8/6.5 & 94.8/55.2 & 51.3/17.6 & \textbf{58.5/20.1} & 65.6/26.9 & 80.6/35.5 & 57.0/27.0\\ \cdashline{2-16}

& W-S & 0.24 & 68.9/31.8 & 49.5/25.7 & 84.8/55.4 & 228.6/164.5 & 33.4/10.3 & 129.15/87.85 & 75.25/36.55 & 79.73/39.3 & 103.83/63 & 112.69/70.69 & 96.6/58.5 \\ 

& W-M & 0.77 & 55.1/24.2 & 37.6/19.6 & 71.5/43.7 & 129.7/89.4 & 24.0/7.1 & 103.9/71.4 & 59.0/23.9 & 66.8/27.6 & 90.7/55.7 & 95.2/56.2 & 73.4/41.9 \\

& W-L-v2 & 1.5 & 46.9/19.6 & 33.7/16.9 & 60.6/37.7 & 101.1/71.1 & 19.7/5.6 & 106.9/74.6 & 51.2/19.6 & 60.2/25.2 & 73.2/41.2 & 86.9/50.1 & 67.4/38.3 \\

& W-L-v3 & 1.5 & 43.2/16.9 & \textbf{20.4/8.6} & 44.6/22.5 & 82.0/47.7 & \textbf{16.4/4.8} & 103.8/68.9 & 52.7/18.9 & 64.3/26.4 & 72.3/35.9 & 86.0/43.3 & 58.6/29.4 \\ \cline{2-16}

& DW-8-8 & 0.44 & 55.0/23.2 & 44.4/19.2 & 69.2/40.4 & 91.0/55.5 & 36.1/13.3 & 91.5/49.6 & 71.4/31.2 & 78.4/35.6 & 82.5/41.2 & 87.5/44.9 & 70.7/35.4 \\

& DW-16-16 & 0.80 & 48.0/18.9 & 33.2/12.5 & 57.1/29.6 & 84.1/46.2 & 26.2/8.5 & 83.8/40.2 & 57.8/20.5 & 68.2/26.2 & 72.0/31.0 & 80.0/35.6 & 61.0/26.9 \\

& DW-32-16 & 1.12 & 45.6/17.7 & 27.7/10.3 & 51.2/26.1 & 80.9/\textbf{43.4} & 22.0/6.6 & \textbf{80.5/35.1} & 52.6/17.1 & 62.9/22.4 & 66.7/26.3 & 77.3/32.6 & 56.7/23.8 \\

& DW-16-32 & 1.22 & 47.4/18.8 & 29.9/11.8 & 55.5/30.7 & 84.7/50.2 & 26.0/7.9 & 84.8/45.7 & 57.4/20.4 & 67.4/28.2 & 71.7/32.3 & 81.5/40.9 & 60.6/28.7 \\
\cdashline{2-16}

& DW-16-16++ & 0.80 & 44.1/17.1 & 28.5/10.5 & 54.5/28.5 & 83.2/45.6 & 22.4/6.9 & 82.3/38.7 & 55.4/18.9 & 65.2/24.9 & 69.3/28.2 & 76.8/33.0 & 58.2/25.2\\
& DW-32-16++ & 1.12 & 44.7/17.3 & 25.2/10.0 & 48.8/25.2 & \textbf{79.0}/43.7 & 20.2/5.0 & 76.4/35.4 & \textbf{50.0/15.9} & 60.1/21.8 & \textbf{63.2/24.7} & \textbf{73.5/31.5} & \textbf{54.1/23.1} \\

\bottomrule 
\end{tabular}
}
\caption{ \label{main-results}
WER/CER scores after normalization and removing diacritics as well as on orthographic transcription. Average is the mean score across all the evaluation sets. All distilled models are trained with a filtering threshold of 80. We report the score on the test split of each dataset.
Abbreviations. \textbf{W} - Whisper, \textbf{FT} - Finetuned, \textbf{M} - Medium, \textbf{L} - Large, \textbf{S} - Small, \textbf{D} - Distil.
}
\end{table*}

We evaluate all models on four versions of CV (6.1, 9.0, 11.0, 15.0), MGB-2, MGB-3, MGB-5, FLEURS, and our five novel dialectal sets. We report WER and CER scores on the orthographic and normalized predictions (as per Section~\ref{text-preprocessing}) in Table~\ref{main-results} for \textit{test} splits and in Table~\ref{validation-all} (Appendix~\ref{results-appendix}) for \textit{dev} splits. CV15.0 results are included in Table~\ref{main-results} and other versions can be found in Appendix~\ref{results-appendix} Table~\ref{rest-cv-versions}.

\noindent\textbf{Commercial Systems and Supervised Models.} The supervised finetuned (SFT) baselines are trained on MGB-2, MGB-3, and the CV datasets. Other evaluation sets thus represent out-of-distribution data. As a result, we see that supervised HuBERT (15.4) and Whisper (25.2) outperform all other models on in-distribution data MGB-2 and MGB-3, respectively. However, these baselines often perform poorly on all other evaluation sets that are not in their training data. On our private in-house data, the supervised models usually produce more incorrect words than the number of words in the corresponding reference. We find varying levels of transcription difficulty for these models when evaluated on distinct dialects and linguistic varieties. 


The \textit{Amazon transcribe} system performs well on our in-house data compared to the supervised baselines. It gives $45.5$\% WER on JOR which is not too far from the best WER of $41.5$\% by \textit{SeamlessM4T-large-v2}. We find that it struggles with ALG, which goes along the trend noticed with all models.

\noindent\textbf{Zero-Shot Models.}
We find that both Whisper and SeamlessM4T models perform quite well on CV\footnote{We refer to CV15.0 as CV.} and FLEURS in zero-shot setting. More specifically, the best Whisper model shows WER scores of 15.8\% and 11.3\% on CV and FLEURS, respectively, while the best SeamlessM4T achieves 9.7\% and 7.6\% WER.
MMS shows the lowest performance of the zero-shot models and the second lowest across all model types with an 82.5\% average WER. 
Meanwhile, a consistent challenge across all models is observed with MGB-5, followed by MGB-3. These last two datasets involve dialects; namely, EGY and MOR dialects respectively. The transition from MSA to dialects thus marks a significant drop in performance, indicating the models' difficulties in adapting to dialectal variations.
This pattern becomes even more apparent when looking at the results of our in-house data. All models particularly struggle with the ALG and YEM dialects, whereas JOR and PAL are less challenging to transcribe. This underscores the distinct issues that dialectal diversity poses to current ASR systems.
The best-performing model overall on both the existing datasets and our new data is \textit{SeamlessM4T-large-v2}, showing a significant improvement in performance compared to its previous version. Although the size is the same between the two systems, \citet{m4t} attributes the higher performance to its novel \textit{UnitY2} architecture.
We also find that both the SeamlessM4T and Whisper family models consistently improve as we increase in size, except for \textit{SeamlessM4T-medium} (48.1\% WER) which outperforms \textit{SeamlessM4T-large-v1} (51.1\% WER) model on average.

\noindent\textbf{Distilled Models.}\footnote{We call the models trained on 500K segments DW-16-16++ and DW-32-16++, and the model trained on 1M segments DW-16-16-1M.}
We distill a wide range of models of varying sizes from \textit{Whisper-large-v2} by reducing the number of encoder and decoder blocks. Our smallest distilled model, which has eight encoder and decoder blocks (resulting in approximately a 75\% reduction in parameters from the teacher model), outperforms \textit{Whisper-medium} with a WER of 64.8\% compared to 65.4\%, while being half the size (Table~\ref{main-results}). When comparing our distilled models with smaller Whisper variants, we find that DW-16-16 outperforms \textit{Whisper-medium} by over 12 points. However, both these models are similar in size.

As expected, we observe that increasing the number of layers in the distilled model enhances its performance. Consequently, our best-performing distilled model, DW-32-16++ (WER 45.0\%), surpasses all other models, including \textit{Whisper-large-v3} (WER 49.5\%) and \textit{SeamlessM4T-v2} (WER 47.0\%), despite being half of its size (see Table~\ref{main-results}).


To sum up, our best-performing distilled models yield the best results in terms of WER on four out of ten evaluated datasets and are on par with an overall best model in terms of average WER while being half in size. However, when looking at the average performance across in-house data only, it outperforms all other systems with a $56.9$\% WER, whereas the best zero-shot model (Seamlessm4T-large-v2) has $61.74$\% WER and teacher model (Whisper-large-v2) $67.6$\%, showing substantial improvement on unseen dialects. We report the average across benchmark and in-house data in Appendix~\ref{results-appendix} Table \ref{tab:average-category-table}. 

Our results underscore the distilled models' inherent efficiency and generalization to unseen dialects, possibly resulting from the mixture of data. This may imply that the process retains the critical linguistic and acoustic features necessary for high-quality ASR in a linguistically diverse setting.

\noindent\textbf{Orthographic, Normalized, and Non-Diacritized Evaluation.} To better understand the effect of normalization and diacritics removal, we calculate WER/CER on \textit{orthographic}, \textit{normalized}, and \textit{normalized+non-diacritized} (ND) transcriptions of \textit{Whisper-large-v2}. We report the results in Table~\ref{pre-v-no-pre}. With normalization, the WER goes down on CV from $47.1$\% to $39.4$\%. Similarly, we see a near $50$\% drop in WER on FLEURS which suggests that the model is much more prone to miss diacritics than missing entire words. However, in the case of MGB-3 and MGB-5, we do not notice any significant changes after pre-processing, which again shows the poor generalization capability of Whisper on unseen and linguistically diverse data.  
\begin{table}[h!]
\begin{tabular}{lccc}
\toprule
Dataset      & \multicolumn{1}{l}{\begin{tabular}[c]{@{}l@{}}Ortho\\ \end{tabular}} & \multicolumn{1}{l}{Norm} & \multicolumn{1}{l}{\begin{tabular}[c]{@{}l@{}}Norm + ND\end{tabular}} \\ \midrule
CV & 47.1/18.9                                                                      & 39.4/17.0                      & 19.4/6.8                                                                                   \\
MGB3         & 52.4/28.2                                                                      & 46.6/23.9                      & 43.5/21.9                                                                                  \\
MGB5         & 85.2/52.2                                                                      & 83.6/49.5                      & 83.0/49.1                                                                                  \\
FLEURS       & 20.3/5.9                                                                       & 17.4/5.0                       & 11.6/3.7   \\  \bottomrule                                                                              
\end{tabular}

\caption{\label{pre-v-no-pre}
WER/CER scores on orthogonal, normalized, and without diacritics outputs produced by \textit{Whisper-large-v2}.  Abbreviations: \textbf{Norm} - Normalized, \textbf{ND} - No Diacritics.
}
\end{table}

\noindent\textbf{Effect of WER Threshold.\label{subsec:threshold}}
The knowledge distillation framework that we follow~\cite{gandhi2023distilwhisper} involves pseudo-labels filtered by WER. While we initially use a WER threshold of 80 in Table~\ref{main-results}, we experiment with different values to find an optimal threshold value that yields better results across different evaluation sets while reducing the amount of data required, subsequently resulting in faster training. The summary of our results is illustrated in Figure~\ref{fig:lambda-experiments} and the detailed results can be found in Table~\ref{tab:lambda-experiments-all} in Appendix~\ref{results-appendix}. From our experiments with the DW-16-16 and DW-32-16 models (trained on 100K segments), we find that discarding examples where the WER is above 80\% (amounting to about 28\% of the total examples) results in the best overall performance across different evaluation setups closely followed by the 20\% WER threshold. Both these models significantly outperform the base teacher model \textit{whisper-large-v2} and the \textit{whisper-medium model}, which is comparable in size. That being said, reducing the threshold from 20\% to 10\% worsens the models' performance. However, training the models without applying any filtering still outperforms the zero-shot baselines. Based on these results, we conclude that there exists a trade-off between the quality and quantity of the distillation data. This implies that we can distill small and compute-efficient language-specific speech recognition models without training on any labeled speech data while being on par or better than the base models.
\begin{figure}[h!]
    \centering
    \includegraphics[width=\linewidth]{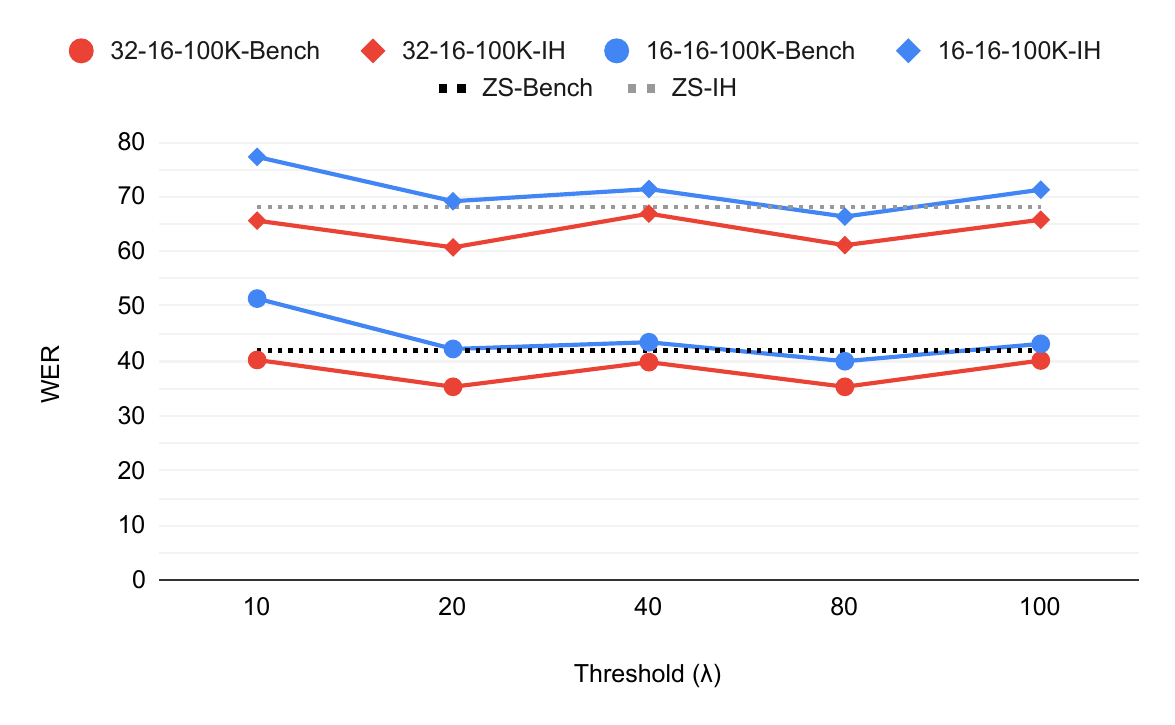}
    \caption{\label{fig:lambda-experiments}
    Average WER on five MSA benchmarks and five dialects from our in-house data with different filtering thresholds. The dotted flat line represents the \textit{Whisper-large-v2} (teacher) in the zero-shot setting. Abbreviations: \textbf{Bench} - Benchmark. \textbf{IH} - In-house. \textbf{ZS} - Zero-shot.}
    \label{fig:enter-label}
\end{figure}

\noindent\textbf{Data Scaling.}
We train all of our models on 100K speech segments ($\approx$100 hours) sampled from the mixture of over 3M segments ($\approx$4000 hours)  described in Section~\ref{datasets}. We increase the data size from 100K to 500K and then up to 1M segments to study the effect of the quantity of the data. With the filtering threshold set to 20\%, DW-16-16 trained on 500K segments outperforms the zero-shot teacher baseline on the MSA benchmark (36.7\% WER compared to 42.0\%) and is significantly better on the in-house data. This trend remains consistent after scaling the data to 1M segments: the model achieves 35.0\% and 60.0\% WER on the MSA benchmark and in-house data, respectively, compared to 42.0\% and 68.0\% from the zero-shot baseline, despite being half its size. 
\begin{figure}[h!]
    \includegraphics[width=\linewidth]{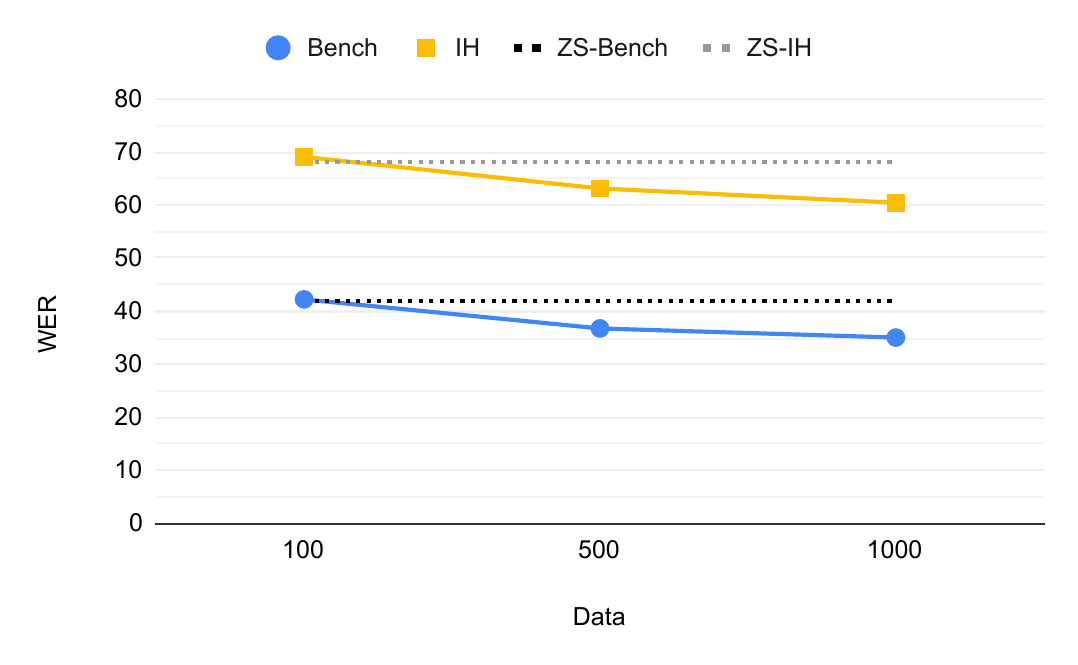}
    \caption{\label{fig:scaling}
    Average WER from DW-16-16 model trained with different amounts of data. The dotted line represents the \textit{Whisper-large-v2} zero-shot baseline. Abbreviations: \textbf{Bench} - Benchmark. \textbf{IH} - In-house. \textbf{ZS} - Zero-shot.
    }
    \label{fig:enter-label}
\end{figure}


\section{Error Analysis}\label{error-analysis}

To gain a better understanding of the results, we conduct an error analysis on our in-house data by randomly sampling $20$ sentences per dialect from each models' outputs, with the aim of identifying the specific types of errors present. We then categorize the errors into the following types:

\noindent\textit{\textbf{MSA Translation}}: The transcription is semantically accurate but employs words in MSA that differ from the dialectal words spoken in the utterance.

\noindent\textit{\textbf{Hallucination}}: The transcription is found to be both semantically and acoustically distant from the utterance.

\noindent \textit{\textbf{Deterioration}}: The transcription is either gibberish (random characters) or involves an excessive repetition of the same word or expression. 

\noindent \textit{\textbf{Incomplete Transcription}}: Parts of the utterance are omitted and do not appear in the transcription.

\noindent \textit{\textbf{Empty Transcription}}: The model fails to generate a prediction at all.

\noindent \textit{\textbf{Dialectal Inaccuracies}}: The prediction and ground truth mismatch is of dialectal nature. Instances such as unrecognized dialectal words, first names, cultural expressions, pronunciations (e.g. Emirati Arabic subbing \textit{ya} {\small\<y>} for \textit{jim} {\small\<ج>}) and alternate orthographies fall in this category.

An example for each of these categories can be found in Table \ref{tab:error-analysis-examples} in Appendix \ref{error-analysis-appendix}.

Upon inspecting the initial results, we decided to further analyze the performance discrepancies among the ASR models by looking at the most problematic transcriptions. Due to the tedious aspect of this exercise, we limit this part of the analysis to five models: the best supervised baseline (HuBERT), the best Whisper (W-L-v3), the best SeamlessM4T (SM4Tv2), the best distilled (DW-32-16), and finally Whisper-medium (W-M) (given its closeness in size to DW-32-16). We set a threshold of $75\%$ CER and look at all the transcriptions with a higher error rate. It is noteworthy that a transcription could embody multiple error categories simultaneously, such as being both incomplete and translated to MSA.


Our results show that the supervised baselines struggle the most, with W-M amounting to $635$ highly erroneous transcriptions with hallucination (closely followed by deterioration) being the category with the most instances. Our D-W-$32$-$16$ model has the least issues with $108$ cases, most of which are simple inaccuracies. This indicates that this model produces the most \textit{coherent} outputs. In other words, this model is more likely to make predictions that maintain relevance, are logically consistent and closely aligned with the input speech.

The fine-tuned model, HuBERT, makes considerably fewer errors than the supervised baselines but still struggles with a lot of deterioration. This category, however, looks different on HuBERT cases than it does on the Whisper and Seamless M4T models: instead of repeating words or characters, it outputs seemingly random sequences of characters occasionally including a single square bracket. These characters are stringed together in word-sized sequences and can include ta marbouta \<ة> in the middle of the ``gibberish'' words (e.g. \<صيةةكو>). It also tends to eliminate spaces between correctly predicted words or fusing two or more half-words together. While empty transcriptions seems to be the category with the least appearances across all models, HuBERT shows a notable increase in these compared to the other systems.
That being said, all models except for HuBERT show MSA Translation errors, with the least observed in the distilled model and the most committed by W-L-v3. We theorize that these could be due to the models being trained on data that include Arabic shows or movies spoken in dialect but mapped with MSA subtitles.
Among the hallucinations of the Whisper models, we also notice a commonly occurring transcription: \<اشتركوا في القناة> (Eng. \textit{subscribe to the channel}), which we believe could be resulting from training models on videos from platforms like YouTube. These videos can include captions that contain these sentences in interludes when the audio contains no speech (noise or background music).
At the dialect level, YEM and UAE are the most problematic across all models, surprisingly exceeding the ALG dialect given its higher error rate overall. PAL and Jordanian are the least challenging, which goes in line with the systems' overall performance on them.
The exact statistics are provided in Table \ref{tab:error-stats} in Appendix \ref{error-analysis-appendix}.

\section{Conclusion}\label{conclusion}
We present a comprehensive evaluation of multilingual ASR systems on a wide range of Arabic varieties and dialects to assess the robustness and generalization capability of these systems to linguistic variations. We then distill small dedicated models for Arabic ASR from large multilingual speech models (Whisper). We evaluate our distilled models on ten diverse datasets and find that despite being 25-50\% smaller, they outperform the base model and are on par with state-of-the-art models twice their size. We also find our distilled models to be the most robust to linguistic diversity. We further conduct a comprehensive error analysis to investigate the nature of the errors these models make. We find that speech models with language model decoding are more prone to hallucination compared to other models. Our work reveals an inherent limitation of these models to generalize beyond their training data. In the future, we intend to expand this work to low-resource and unseen languages.
\section{Limitations}\label{limitations}

In this study, we distill small Whisper models from relatively large ones via pseudo-labeling. While our distilled models are compute efficient and maintain a performance similar to or better than the base teacher model, we believe that our work has several limitations which we outline below.

\noindent\textbf{Evaluation.} Arabic is a linguistically rich and complex language with over 400 million speakers~\cite{400m}, resulting in its wide range of varieties and dialects. We evaluate all the models on ten different datasets representing different varieties, including five novel dialects collected and curated by native speakers and never seen before by any models. However, our varieties do not cover all Arabic-speaking regions. We aim to address this in future work by covering more varieties and dialects.

\noindent\textbf{Efficiency.}
Our distilled models are 25-75\% compute efficient while maintaining the same performance as big models. However, the training process demands substantial computational resources. Our rough approximation indicates an expenditure of more than 3000 A100 (80G) GPU hours in our experiments, equivalent to over 500 kg of CO2 emissions of which zero percent is directly offset. To offer perspective, this carbon output aligns with what a typical internal combustion engine emits during a distance of about 2,000 kilometers. Our estimations rely on the \href{https://mlco2.github.io/impact#compute}{Machine Learning Impact calculator} presented in \cite{lacoste2019quantifying}.

\noindent\textbf{Distillation Training Data.}
We distilled four variants of student models using 100K and 500K segments of which approximately 25\% are filtered. We see improvement going from 100K ($\approx$$100$ hours) to 500K ($\approx$$500$ hours) segments. As \cite{gandhi2023distilwhisper} shows going over 1000 hours results in a better model, we aim to study how distillation can be done under a low resource setting which is why we do not scale the data. Additionally, we also keep the WER threshold high (80) so that we remain close to a setting where no labeled data is available (even for filtering). It would be interesting, however, to see how distilled models may perform on unfiltered data in low-resource setting.  

\noindent\textbf{Nature of Speech Data.} Despite putting together a never-seen dataset of under-represented Arabic dialects, we realize that sourcing our data from television series renders its nature distant from speech spoken \textit{in the wild}. This type of content tends to be more ``theatrical'' and involves different elements such as background music and laughing tracks that do not accurately reflect regular conversational Arabic. Consequently, this could fail to accurately portray the performance of these models on real speech.

\section{Ethics Statement}\label{ethics-statement}
\label{sec:ethics}

\textbf{Data Collection and Release.} Given that we collect our data from TV series available on YouTube, we ensure that our use of this data aligns with the principles of fair use, given its application to a non-commercial academic setting. Each annotator of the data was made fully aware of the research objectives of the study and the intended use of their annotations.

\noindent\textbf{Intended Use.} We believe our work will embolden further research on distilling small and efficient models from large and powerful foundation models especially applied to medium and low-resource languages. Our results show that small distilled models can yield on-par performance on even better results compared to large teacher models. Therefore, our work can raise the interest among the researchers who work on developing efficient machine learning systems under low resource settings however crucial to a wide range of population.
\\
\noindent \textbf{Potential Misuse and Bias.}
Our distilled models can efficiently generate high-quality transcripts for multiple Arabic dialects and have the potential to be misused. Since there exists little-to-no clarity on the nature of the training data of the teacher model, our distilled models can produce potentially harmful and biased content that they can inherit from the teacher model. In addition to that, in our human evaluation, we find that these are susceptible to generating examples from the training data which raises the threat of information leakage. Therefore, we recommend against our distilled models being used without a careful prior consideration of potential misuse and bias.

\section*{Acknowledgments}\label{sec:acknow}
We acknowledge support from Canada Research Chairs (CRC), the Natural Sciences and Engineering Research Council of Canada (NSERC; RGPIN-2018-04267), the Social Sciences and Humanities Research Council of Canada (SSHRC; 895-2020-1004; 895-2021-1008), Canadian Foundation for Innovation (CFI; 37771), Digital Research Alliance of Canada,\footnote{\href{https://alliancecan.ca}{https://alliancecan.ca}} and UBC Advanced Research Computing-Sockeye.\footnote{\href{https://arc.ubc.ca/ubc-arc-sockeye}{https://arc.ubc.ca/ubc-arc-sockeye}}

\bibliography{anthology,custom}
\bibliographystyle{acl_natbib}

\appendix

\section{Related Work}\label{related-work-appendix}
While early ASR systems were primarily hybrid~\cite{early-hybrid}, often in the form of combinations of Hidden Markov Models (HMMs) and either Gaussian Mixture Models (GMMs) or Deep Neural Networks (DNNs), the desire for simpler architectures led to a shift towards End-to-End (E2E) models \cite{e2e}. This was made possible in part thanks to the availability of extensive labeled datasets and increased computational power. Transformers \cite{transformer} have come to light as the dominant architecture in modern ASR systems \cite{sru}, owing to their attention mechanism's ability to model long-range dependencies all while being scalable and efficient.

OpenAI's Whisper \cite{whisper}, a weakly supervised encoder-decoder Transformer, was trained on an extensive 630K hours of multilingual data, 739 of which are in Arabic. Whisper supports multilingual ASR, Automatic Speech Translation (AST) to English, and Language Identification (LID).
Massively Multilingual Speech (MMS) \cite{pratap2023scaling}, a system for multilingual ASR, speech synthesis (TTS) and LID build by Meta, is the result of pre-training wav2vec 2.0 \cite{wav2vec2} models (300M and 1B parameter versions) on 419K hours from 6 different corpora, spanning 1406 languages. For the ASR task, they fine-tune the pre-trained 1B model on 44.7K hours of labeled data in 1107 languages using Connectionist Temporal Classification (CTC) \cite{ctc}.
Meta also developed SeamlessM4T v2 \cite{m4t, Communication2023SeamlessM4TMM}, a collection of models featuring the new w2v-BERT 2.0 speech encoder pre-trained on 4.5M unlabeled data hours and fine-tuned on automatically aligned pairs. It supports 100 languages and its Arabic training data includes 119K hours of raw audio and 822 hours of labeled data.
Another system that performs ASR and AST is Google's Universal Speech Model (USM) \cite{usm}, a 2B parameter model employing a Conformer encoder. It was pre-trained using BEST-RQ \href{https://arxiv.org/pdf/2202.01855.pdf}{\cite{best-rq}} on 12M unlabeled hours covering 300 distinct languages. Supervised ASR training was then used on the Conformer features using either CTC or Listen, Attend and Spell (LAS) \cite{las} transducers using 90K hours of labeled data across 70 languages.
XLS-R \cite{xls-r} is yet another wav2vec 2.0-based model used for ASR, AST and speech classification tasks (LID and Speaker ID). It comes in variants of 0.3B, 1B and 2B parameters, trained on 436K hours (95 are in Arabic) that include 128 languages.
ArTST \cite{artst} is a SpeechT5 model focused on MSA and fine-tuned on the MGB3 dataset for ASR and on ASC \cite{asc} and ClArTTS \cite{clartts} for TTS.

\section{Training}\label{training}
\begin{table*}[]
\centering
\begin{tabular}{ll}
\toprule
Parameter                     & Value                  \\
\midrule
$warmup\_steps$                 & $50$                     \\
$learning\_rate$                & $0.0001$                 \\
$lr\_scheduler\_type$           & $constant\_with\_warmup$ \\
$batch\_size$                   & $128$                    \\
$max\_label\_length$            & $225$                    \\
$gradient\_accumulation\_steps$ & $1$                      \\
$dtype$                         & $bfloat16$      \\

\bottomrule
\end{tabular}
\caption{\label{training-parameters} Training parameters. We use all the default training parameters provided in Huggingface Seq2SeqTrainingArguments unless otherwise specific in this table.  
}
\end{table*}

\section{Results}\label{results-appendix}
\begin{table*}[h!]
\centering
\Large 
\renewcommand{\arraystretch}{1.2}   
\resizebox{1.\linewidth}{!}{
\begin{tabular}{lccccc|ccccc}
\toprule
\multirow{2}{*}{Model} & \multicolumn{5}{c}{Orthographic} & \multicolumn{5}{c}{Normalized + No Diacritics}\\ 
 & CV15.0 & MGB2 & MGB3 & MGB5 & FLEURS & CV15.0 & MGB2 & MGB3 & MGB5 & FLEURS \\ 
\midrule
XLS-R & 91.3/45.5 & 98.0/58.2 & 99.1/63.7 & 99.8/73.5 & 95.3/46.3 & 85.7/34.2 & 97.7/55.6 & 98.7/61.1 & 99.8/71.8 & 94.8/44.8\\

HuBERT & 79.9/34.4 & 70.0/27.7 & 48.8/18.6 & 98.4/52.7 & 49.4/14.3 & 53.0/17.6 & 55.1/22.5 & \textbf{33.2/12.5} & 96.9/50.2 & 33.9/10.7\\

W-FT & 73.8/36.1 & 50.3/22.8 & 62.8/33.2 & 118.9/75.8 & 30.7/7.6 & 30.1/18.9 & 27.6/17.1 & 52.0/28.4 & 116.7/74.8 & 10.4/3.7\\ \midrule

MMS-all & 107.6/81.2 & 55.5/23.5 & 76.1/35.9 & 94.2/52.0 & 28.7/7.1 & 108.0/78.3 & 48.3/20.0 & 73.8/32.7 & 93.7/49.8 & 22.8/5.6\\ \hdashline

SM4T-M & 48.0/21.7 & 44.9/21.0 & 49.6/25.1 & 92.1/55.2 & 20.4/6.1 & 13.3/4.6 & 31.2/16.2 & 41.7/20.5 & 88.0/51.5 & 9.2/3.7\\

SM4T-L-v1 & 48.1/21.4 & 44.4/21.4 & 50.6/25.3 & 95.7/60.0 & 22.9/7.5 & 16.5/5.9 & 33.0/17.3 & 43.2/20.9 & 93.2/56.8 & 11.4/5.1\\

SM4T-L-v2 & \textbf{40.0/16.9} & 42.5/20.6 & 46.0/22.6 & 95.9/60.7 & 20.2/6.4 & \textbf{8.3/2.5} & 30.3/16.5 & 36.1/17.7 & 90.8/56.2 & \textbf{7.9}/3.8\\ \hdashline 

W-S & 64.2/29.1 & 83.8/50.6 & 85.4/55.5 & 198.3/140.1 & 36.3/12.1 & 44.0/19.1 & 75.2/47.5 & 82.0/51.3 & 197.2/140.0 & 30.4/10.6\\

W-M & 59.1/26.7 & 65.4/35.8 & 70.7/43.8 & 145.1/105.3 & 24.2/6.5 & 23.7/9.7 & 52.7/32.1 & 63.2/39.6 & 143.2/104.1 & 15.5/4.5\\ 

W-L-v2 & 53.2/23.4 & 57.0/30.0 & 58.4/34.9 & 118.4/86.0 & 18.6/4.8 & 15.4/5.5 & 41.4/25.6 & 50.5/30.5 & 116.5/84.1 & 10.2/2.9\\

W-L-v3 & 50.3/21.8 & \textbf{37.5/17.1} & \textbf{44.1/20.6} & 88.1/53.5 & \textbf{17.1/4.2} & 12.2/3.9 & \textbf{23.4/12.7} & 34.7/15.7 & 86.0/50.8 & 8.9/\textbf{2.6}\\ \midrule

DW-8-8 & 59.1/26.2 & 59.5/29.3 & 68.5/39.2 & 94.9/60.1 & 38.1/14.2 & 27.2/9.4 & 48.5/25.3 & 64.1/35.8 & 94.0/57.9 & 31.3/12.2\\

DW-16-16 & 53.5/23.2 & 50.5/22.4 & 56.2/27.9 & 89.8/51.2 & 27.4/8.6 & 17.5/5.4 & 35.9/17.7 & 50.0/23.9 & 88.4/48.7 & 19.5/6.7\\

DW-32-16 & 51.9/22.3 & 45.8/20.1 & 50.4/24.3 & 87.4/47.7 & 23.1/6.7 & 14.7/4.4 & 30.3/15.3 & 43.3/20.0 & 85.6/45.1 & 14.9/4.9\\

DW-16-32 & 52.9/22.9 & 48.1/23.3 & 55.4/29.8 & 90.4/54.0 & 24.8/9.0 & 16.9/5.4 & 34.9/19.0 & 49.2/25.8 & 88.9/51.6 & 17.4/7.1\\  

\hdashline
DW-16-16++ & 52.4/22.5 & 47.3/21.7 & 53.3/27.7 & 87.0/50.1 & 23.4/6.6 & 14.9/4.6 & 32.5/17.0 & 46.2/23.4 & 85.3/47.5 & 15.0/4.7\\

DW-32-16++ & 51.4/22.0 & 41.4/18.1 & 48.1/24.1 & \textbf{85.0/47.3} & 19.2/5.6 & 13.2/3.9 & 27.2/13.7 & 40.3/19.7 & \textbf{82.7/44.6} & 11.1/3.7 \\


\bottomrule
\end{tabular}
}
\caption{ \label{validation-all}
WER/CER on validation split of each dataset. Our in-house data only includes a single split reported in Table~\ref{main-results}.
Abbreviations. W - Whisper, FT - Finetuned, M - Medium, L - Large, S - Small, D - Distil.
}
\end{table*}

\begin{table*}[h!]
\centering
\Large 
\renewcommand{\arraystretch}{1.}   
\resizebox{1.\linewidth}{!}{
\begin{tabular}{clccc|ccc}

\toprule
\multirow{2}{*}{Split} & \multirow{2}{*}{Model} & \multicolumn{3}{c}{Orthographic} & \multicolumn{3}{c}{Normalized + No Diacritics}\\
& & CV6.1 & CV9.0 & CV11.0 & CV6.1 & CV9.0 & CV11.0 \\ 
\midrule
\multirow{17}{*}{Test} & XLS-R & 92.2/47.4 & 92.9/47.1 & 92.8/46.9 & 88.0/37.7 & 89.9/39.6 & 89.8/39.5\\

& HuBERT & 78.9/33.1 & 76.6/31.1 & 76.5/31.0 & 52.0/17.8 & 54.7/18.7 & 54.8/18.8\\

& W-FT & 74.9/36.7 & 69.8/33.5 & 69.5/33.3 & 32.8/21.8 & 34.9/21.1 & 35.0/21.1\\ \cline{2-8}

& MMS-all & 106.1/82.4 & 106.0/82.6 & 105.9/82.5 & 106.8/80.2 & 106.5/80.9 & 106.4/80.9\\ \cdashline{2-8}

& SM4T-M & 40.8/17.4 & 42.1/18.2 & 42.1/18.1 & 13.2/4.9 & 16.2/5.7 & 16.2/5.7\\

& SM4T-L-v1 & 43.3/19.2 & 44.2/19.2 & 44.0/19.0 & 15.8/6.4 & 19.6/7.4 & 19.6/7.3\\

& SM4T-L-v2 & \textbf{34.2/13.5} & \textbf{37.5/15.8} & \textbf{37.4/15.7} & \textbf{8.4/2.8} & \textbf{11.1/3.5} & \textbf{11.1/3.5}\\ \cdashline{2-8} 

& W-S & 73.9/35.4 & 68.7/31.7 & 68.9/31.9 & 44.0/19.2 & 40.3/16.3 & 40.3/16.4\\

& W-M & 59.1/26.3 & 55.4/24.4 & 55.5/24.6 & 25.8/11.9 & 29.5/13.0 & 29.8/13.4\\ 

& W-L-v2 & 51.4/21.9 & 47.9/20.3 & 47.7/20.2 & 16.2/7.0 & 19.8/8.2 & 19.9/8.2\\

& W-L-v3 & 49.2/19.8 & 43.7/17.3 & 43.6/17.1 & 12.8/4.4 & 15.5/5.1 & 15.6/5.2\\ \cline{2-8}

& DW-8-8 & 58.2/24.8 & 55.4/23.5 & 55.2/23.3 & 28.5/10.4 & 32.5/12.2 & 32.6/12.2 \\

& DW-16-16 & 52.6/21.3 & 48.5/19.3 & 48.3/19.1 & 18.5/6.1 & 21.9/7.2 & 22.1/7.2\\

& DW-32-16 & 50.5/20.2 & 46.2/18.0 & 46.0/17.9 & 15.2/4.8 & 18.5/5.8 & 18.7/5.8\\

& DW-16-32 & 52.0/21.1 & 47.9/19.2 & 47.7/19.0 & 17.6/5.9 & 21.2/7.2 & 21.3/7.3\\ \cdashline{2-8} 

& DW-16-16++ & 51.2/20.6 & 46.6/18.4 & 46.5/18.2 & 15.8/5.2 & 19.0/6.2 & 19.1/6.2\\

& DW-32-16++ & 49.8/19.9 & 45.2/17.7 & 45.0/17.5 & 13.7/4.4 & 16.9/5.4 & 17.0/5.5\\ 

\midrule \midrule

\multirow{17}{*}{Val.} & XLS-R & 92.1/48.6 & 91.1/45.3 & 91.3/45.6 & 86.6/36.2 & 85.4/34.0 & 85.7/34.2\\

& HuBERT & 82.6/36.8 & 79.9/34.3 & 80.0/34.4 & 54.9/18.9 & 53.1/17.7 & 53.0/17.6\\

& W-FT & 81.3/41.4 & 74.3/36.7 & 74.5/36.7 & 36.5/24.1 & 31.1/19.8 & 30.9/19.6\\

\cline{2-8}

& MMS-all & 105.8/81.9 & 107.6/81.3 & 107.6/81.3 & 106.2/78.7 & 107.9/78.3 & 108.0/78.3\\ \cdashline{2-8}

& SM4T-M & 48.9/22.2 & 48.0/21.7 & 48.1/21.8 & 13.5/4.8 & 13.2/4.5 & 13.2/4.5\\

& SM4T-L-v1 & 49.6/22.6 & 48.1/21.6 & 48.4/21.7 & 16.6/6.0 & 16.5/5.9 & 16.6/5.9\\

& SM4T-L-v2 & \textbf{40.7/17.6} & \textbf{40.2/17.2} & \textbf{40.3/17.1} & \textbf{8.2/2.6} & \textbf{8.2/2.5} & \textbf{8.3/2.5}\\ \cdashline{2-8}

& W-S & 67.1/31.2 & 64.9/29.7 & 64.8/29.5 & 40.2/17.8 & 44.1/19.3 & 44.2/19.4\\

& W-M & 64.5/30.2 & 58.7/26.4 & 59.0/26.6 & 27.3/12.1 & 23.6/9.3 & 23.6/9.3\\

& W-L-v2 & 57.2/25.9 & 52.8/23.3 & 53.1/23.4 & 17.2/6.7 & 15.4/5.5 & 15.3/5.4\\ 

& W-L-v3 & 53.9/24.2 & 50.0/21.8 & 50.3/22.0 & 13.4/4.7 & 12.2/4.0 & 12.2/3.9\\ \cline{2-8}

& DW-8-8 & 62.7/28.7 & 58.9/26.2 & 59.1/26.4 & 29.0/10.4 & 27.4/9.4 & 27.3/9.4\\

& DW-16-16 & 57.3/25.5 & 53.3/23.1 & 53.5/23.3 & 19.0/6.2 & 17.5/5.5 & 17.5/5.4\\

& DW-32-16 & 55.5/24.5 & 51.7/22.3 & 51.9/22.4 & 15.9/4.9 & 14.8/4.4 & 14.7/4.4\\

& DW-16-32 & 56.7/25.3 & 52.7/22.9 & 53.0/23.1 & 18.4/6.1 & 17.0/5.4 & 16.9/5.3\\ \cdashline{2-8}

& DW-16-16++ & 55.9/24.6 & 52.1/22.4 & 52.4/22.5 & 15.8/5.0 & 15.1/4.6 & 15.0/4.5\\

& DW-32-16++ & 55.1/24.3 & 51.2/22.1 & 51.4/22.2 & 14.3/4.6 & 13.3/4.0 & 13.2/4.0\\

\bottomrule

\end{tabular}
}
\caption{ \label{rest-cv-versions}
Test and validation split results for other common voice versions.
Abbreviations. W - Whisper, FT - Finetuned, M - Medium, L - Large, S - Small, D - Distil.
}
\end{table*}

\begin{table*}[]
    \centering
    \begin{tabular}{lccc}
    \toprule
         Model & Overall Avg. & Avg. Benchmark & Avg. In-House\\
         \midrule
         Amazon & 61.0/41.8 & -/- & -/- \\
         \midrule
         XLS-R & 97.7/58.4 & 96.1/53.2 & 99.4/63.5\\
         HuBERT & 66.7/27.3 & 51.5/20.4 & 81.9/34.2\\
         W-FT & 67.7/39.9 & 42.2/24.5 & 93.2/55.22\\
         \midrule
         MMS-all & 82.5/54.4 & 66.9/36.2 & 98.0/72.6\\
         \hdashline
         SM4T-M & 48.1/21.7 & 33.9/17.3 & 62.3/26.0\\
         SM4T-L-v1 & 51.7/24.7 & 37.4/19.5 & 66.0/29.9\\
         SM4T-L-v2 & 47.0/22.6 & 32.3/17.7 & 61.7/27.6\\
         \hdashline
         W-S & 80.8/45.7 & 66.0/35.3 & 95.6/56.1\\
         W-M & 65.4/38.5 & 54.3/32.9 & 76.4/44.1\\
         W-L-v2 & 55.1/32.3 & 42.0/25.7 & 68.2/38.9\\
         W-L-v3 & 49.5/25.4 & \textbf{31.4/15.6} & 67.7/35.2\\
         \midrule
         DW-8-8 & 64.8/32.1 & 51.3/26.2 & 78.3/38.0\\
         DW-16-16 & 53.2/23.2 & 40.0/18.6 & 66.3/27.9\\
         DW-32-16 & 48.2/20.0 & 35.4/16.3 & 61.1/23.7\\
         DW-16-32 & 53.0/25.1 & 39.4/19.6 & 66.5/30.6\\
         \hdashline
         DW-16-16++ & 49.5/22.5 & 36.7/17.8 & 62.3/27.2\\
         DW-32-16++ & \textbf{45.0/19.2} & 33.0/15.7 & \textbf{56.9/22.7}\\
         \bottomrule

    \end{tabular}
    \caption{Average WER/CER scores on the benchmark, in-house, and overall data. Avg.: Average.}
    \label{tab:average-category-table}
\end{table*}
\begin{table*}[]
    \centering
    \Large 
    \renewcommand{\arraystretch}{1.1}   
    \resizebox{1.\linewidth}{!}{
    \begin{tabular}{llccccccccccc}
    \toprule
         & \multicolumn{2}{c}{Filtering Threshold ($\lambda$)} & \multicolumn{2}{c}{10 (82.8)} & \multicolumn{2}{c}{20 (74.7)} & \multicolumn{2}{c}{40 (54.5)} & \multicolumn{2}{c}{80 (28.0)} & \multicolumn{2}{c}{None (0.0)}\\
        
         Model & Dataset & Split & Orth. & N+ND  & Orth. & N+ND  & Orth. & N+ND  & Orth. & N+ND  & Orth. & N+ND\\
     \midrule
     \multirow{17}{*}{\centering\rotatebox{90}{DW-32-16}} & \multirow{2}{*}{CV15.0} & Test & 45.4/17.7 & 19.3/6.0 & 43.6/16.8 & \textbf{16.9/5.2} & 46.7/18.3 & 20.8/6.8 & 45.6/17.7 & 18.8/5.9 & 47.2/18.8 & 21.2/7.3\\
     & & Dev & 51.6/22.2 & 14.9/4.4 & 50.6/21.8 & \textbf{13.4/3.9} & 52.8/22.9 & 16.6/5.2 & 51.9/22.3 & 14.7/4.4 & 53.2/23.5 & 16.7/5.8\\
     & \multirow{2}{*}{MGB2} & Test & 30.0/11.3 & 26.0/10.3 & 25.6/9.4 & \textbf{20.8/8.3} & 31.8/12.6 & 26.1/11.3 & 27.7/10.3 & 21.1/8.9 & 29.0/11.8 & 22.8/10.4\\
     & & Dev & 48.5/22.5 & 35.8/18.2 & 43.8/19.7 & \textbf{30.1/15.2} & 49.0/22.3 & 35.4/17.8 & 45.8/20.1 & 30.3/15.3 & 52.4/26.1 & 38.2/21.8\\
     & \multirow{2}{*}{MGB3} & Test & 58.9/31.4 & 53.2/27.1 & 51.3/26.3 & 44.8/21.7 & 56.8/30.3 & 50.4/25.9 & 51.2/26.1 & \textbf{43.8/21.4} & 58.3/35.0 & 51.3/30.8\\
     & & Dev & 57.4/29.1 & 51.8/25.1 & 50.0/23.9 & 43.9/19.8 & 55.9/28.3 & 49.5/24.2 & 50.4/24.3 & \textbf{43.3/20.0} & 58.9/35.2 & 51.9/31.5\\
     & \multirow{2}{*}{MGB5} & Test & 84.2/46.1 & 82.4/43.1 & 81.1/43.2 & 79.3/\textbf{40.1} & 85.2/48.5 & 83.4/45.6 & 80.9/43.4 & \textbf{78.9}/40.4 & 92.5/60.8 & 90.5/58.8\\
     & & Dev & 89.8/50.6 & 88.3/47.9 & 87.3/47.1 & 85.7/\textbf{44.4} & 91.6/53.1 & 90.0/50.6 & 87.4/47.7 & \textbf{85.6}/45.1 & 97.4/67.0 & 95.6/65.3\\
     & \multirow{2}{*}{Fleurs} & Test & 27.1/8.9 & 20.0/7.1 & 22.2/6.8 & 14.5/5.0 & 25.0/8.3 & 18.1/6.6 & 22.0/6.6 & \textbf{14.2/4.8} & 23.5/7.0 & 14.9/4.9\\
     & & Dev & 28.1/8.4 & 20.2/6.5 & 22.8/6.3 & 15.0/\textbf{4.6} & 25.2/7.6 & 18.0/5.9 & 23.1/6.7 & \textbf{14.9}/4.9 & 24.5/7.0 & 15.7/4.9\\
     \cdashline{2-13}
     & Avg. B. & -- & 52.1/24.8 & 41.2/19.6 & 47.8/22.1 & \textbf{36.4/16.8} & 52.0/25.2 & 40.8/20.0 & 48.6/22.5 & 36.6/17.1 & 53.7/29.2 & 41.9/24.2\\
     \cmidrule{2-13}
     & ALG	& -- & 83.5/38.6 & 82.6/36.7 & 80.7/35.8 & 79.6/34.0 & 85.1/42.9 & 84.3/41.2 & 80.5/35.1 & \textbf{79.5/33.4} & 88.0/64.2 & 87.3/63.2\\
& JOR	& -- & 58.1/20.1 & 50.7/17.7 & 52.7/17.1 & 44.8/\textbf{14.6} & 58.5/21.3 & 51.3/18.9 & 52.6/17.1 & \textbf{44.4}/14.7 & 55.3/22.4 & 47.6/20.0\\
& PAL	& -- & 68.2/25.3 & 60.9/22.4 & 63.2/21.9 & 55.5/\textbf{19.0} & 68.0/27.2 & 60.8/24.4 & 62.9/22.4 & \textbf{55.0}/19.5 & 65.5/27.9 & 57.1/24.9\\
& UAE	& -- & 71.0/29.3 & 63.6/25.8 & 66.4/25.6 & \textbf{58.1/22.1} & 72.9/32.7 & 65.6/29.4 & 66.7/26.3 & \textbf{58.1}/22.8 & 72.5/38.8 & 63.9/35.6\\
& YEM	& -- & 78.6/34.0 & 70.3/29.4 & 75.2/30.6 & \textbf{65.6/25.6} & 79.9/36.7 & 72.4/32.6 & 77.3/32.6 & 68.5/28.1 & 80.4/53.8 & 73.0/50.9\\
     \cdashline{2-13}
      & Avg. IH & -- & 71.9/29.5 & 65.6/26.4 & 67.6/26.2 & \textbf{60.7/23.1} & 72.9/32.2 & 66.9/29.3 & 68.0/26.7 & 61.1/23.7 & 72.3/41.4 & 65.8/38.9\\ 
     \bottomrule
     \multirow{17}{*}{\centering\rotatebox{90}{DW-16-16}} & \multirow{2}{*}{CV15.0} & Test & 51.8/20.7 & 28.3/9.6 & 47.5/18.7 & 22.4/7.3 & 49.1/19.4 & 24.3/8.0 & 48.0/18.9 & \textbf{22.1/7.2} & 48.3/19.1 & 22.8/7.6\\
     & & Dev & 56.5/24.6 & 22.9/7.3 & 53.0/23.0 & 17.6/\textbf{5.4} & 54.0/23.6 & 19.2/6.1 & 53.5/23.2 & \textbf{17.5/5.4} & 54.1/23.5 & 18.2/5.7\\
     & \multirow{2}{*}{MGB2} & Test & 43.6/16.9 & 39.7/15.7 & 34.4/12.2 & 27.5/\textbf{10.6} & 35.4/13.9 & 29.9/12.5 & 33.2/12.5 & \textbf{26.0}/10.8 & 34.2/13.0 & 26.1/11.2\\
     & & Dev & 59.4/27.4 & 48.5/23.3 & 52.3/23.5 & 37.8/18.8 & 52.0/24.2 & 39.5/20.0 & 50.5/22.4 & \textbf{35.9/17.7} & 55.5/26.5 & 40.6/21.9\\
     & \multirow{2}{*}{MGB3} & Test & 72.0/38.6 & 68.3/34.7 & 61.0/31.9 & 55.5/27.5 & 60.2/31.7 & 54.2/27.4 & 57.1/29.6 & \textbf{50.5/25.1} & 60.2/33.9 & 54.1/29.7\\
     & & Dev & 71.0/36.9 & 67.3/33.3 & 60.1/30.2 & 54.4/26.2 & 59.3/30.4 & 53.6/26.4 & 56.2/27.9 & \textbf{50.0/23.9} & 60.1/32.1 & 54.0/28.3\\
     & \multirow{2}{*}{MGB5} & Test & 90.2/51.7 & 89.1/49.0 & 86.2/47.8 & 84.7/44.8 & 88.8/50.3 & 87.1/47.6 & 84.1/46.2 & \textbf{82.4/43.3} & 96.8/61.3 & 95.1/59.1\\
     & & Dev & 94.2/56.0 & 93.4/53.7 & 91.7/52.5 & 90.5/49.9 & 94.8/57.1 & 93.5/54.9 & 89.8/51.2 & \textbf{88.4/48.7} & 102.1/65.5 & 100.7/63.6\\
     & \multirow{2}{*}{Fleurs} & Test & 36.6/13.3 & 31.6/11.9 & 28.2/9.7 & 21.0/7.9 & 28.7/9.5 & 21.6/7.8 & 26.2/8.5 & 18.8/6.6 & 24.9/7.9 & \textbf{17.6/6.0}\\
     & & Dev & 36.6/12.7 & 31.4/11.3 & 29.0/9.1 & 21.5/7.3 & 29.9/9.9 & 22.6/8.1 & 27.4/8.6 & 19.5/6.7 & 26.7/7.9 & \textbf{19.3/6.2}\\
     \cdashline{2-13}
      & Avg. B. & -- & 61.2/29.9 & 52.1/25.0 & 54.3/25.9 & 43.3/20.6 & 55.2/27.0 & 44.6/21.9 & 52.6/24.9 & \textbf{41.1/19.5} & 56.3/29.1 & 44.9/23.9\\ 
     \cmidrule{2-13}
     & ALG	& -- & 89.9/46.4 & 89.2/44.9 & 85.7/41.3 & 84.8/39.6 & 87.0/46.8 & 86.8/45.3 & 83.8/40.2 & \textbf{83.0/38.5} & 93.8/53.1 & 93.4/51.6\\
& JOR	& -- & 71.4/29.6 & 66.5/27.5 & 62.3/23.1 & 55.5/20.7 & 63.4/24.3 & 56.9/22.1 & 57.8/20.5 & \textbf{50.4/18.2} & 58.9/22.4 & 51.8/20.2\\
& PAL	& -- & 78.7/34.2 & 73.6/31.5 & 70.9/27.8 & 64.6/25.0 & 72.4/31.3 & 66.2/28.6 & 68.2/26.2 & \textbf{61.0/23.3} & 72.3/30.2 & 64.7/27.3\\
& UAE	& -- & 81.7/39.1 & 77.1/36.2 & 74.9/32.4 & 68.2/29.1 & 76.1/35.9 & 69.8/32.9 & 72.0/31.0 & \textbf{64.6/27.7} & 75.5/37.2 & 68.0/34.0\\
& YEM	& -- & 85.1/41.5 & 79.7/37.9 & 80.1/35.6 & \textbf{72.6/31.3} & 83.3/40.8 & 77.2/37.1 & 80.0/35.6 & 72.7/31.6 & 84.8/42.3 & 78.3/38.8\\

    \cdashline{2-13}
    & Avg. IH & -- & 81.4/38.2 & 77.2/35.6 & 74.8/32.0 & 69.1/29.1 & 76.4/35.8 & 71.4/33.2 & 72.4/30.7 & \textbf{66.3/27.9} & 77.1/37.0 & 71.2/34.4 \\ 
     \bottomrule
     \end{tabular}
     }
    \caption{
        Results for different threshold ($\lambda$) values distilling from 100K segments. The value in the bracket along with $\lambda$ represents the ratio of filtered examples. The average of reported results on test split of benchmarks and in-house data. Orth.: Orthographic. N: Normalized. ND: Non Diacritized.
    }
    \label{tab:lambda-experiments-all}
\end{table*}

\section{Error Analysis}\label{error-analysis-appendix}
\begin{table*}[]
\centering
\renewcommand{\arraystretch}{0.1}   
\resizebox{1.\linewidth}{!}{
    \begin{tabular}{lcrrc} 
        \toprule
        Category & Dia. & \makecell{Reference} & \makecell{Prediction} & Model \\
        \midrule
        MSA Trans. & ALG & 
        \begin{tabular}{r}
             {\small \<كلي شغل خرجتلي> \textcolor{Red3}{\<حاسة>} \<أني>} \\
             {\small \<نروح> \textcolor{Red3}{\<باغية>} \<تكميشا راني>}
        \end{tabular} & 
        \begin{tabular}{r}
             {\small \<بأنه يجب أن أخرج من> \textcolor{Red3}{\<أشعر>} \<أنا>}\\
             
             {\small \<أن أقوم> \textcolor{Red3}{\<أريد>} \<الكميش لأنني>}\\
        \end{tabular} & W-M \\
        
        Hallucination  & JOR &
        \begin{tabular}{r}
            {\small \<شو؟ نستسلم يعني ؟ نرضى >}  \\
            {\small \<بوجودها بينا؟>} 
        \end{tabular} &
        \begin{tabular}{r}
            {\small \<نحن نستطيع الوصول إلى الوصول>}  \\
            {\small \<إلى المنزل>}
        \end{tabular} & DW-32-16\\

        Deterioration & PAL & 
        \begin{tabular}{r}
          {\small \<مزبوط قايلة لنجم بدك سلامة؟>}\\
        \end{tabular} &
        \begin{tabular}{r}
          {\small \<اظلو كي ل ت متسالرة يا مان>} \\
        \end{tabular} & HuBERT \\
        
        Incomplete & YEM & 
        \begin{tabular}{r}
          {\small \<اخي كريم، الزهايمر يالله، صلي>}\\
          {\small \<على رسول الله>}
        \end{tabular}
        & {\small \<أخي كري>}& W-M \\

        Unr. Word & ALG &
        \begin{tabular}{r}
            {\small \<دار ، دار والو> \textcolor{Red3}{\<حوست>} \<مولاي>}\\
            {\small \<تقول ما صبت حتى واحد>}
        \end{tabular} &
        \begin{tabular}{r}
            {\small \<جدار دار، والو> \textcolor{Red3}{\<هوا>} \<مولاي،>}\\
            {\small \<تقول ما صبت حتى واحد>}
        \end{tabular} & M4T v1\\

        Unr. Name & JOR &
        \begin{tabular}{r}
            {\small \<إنه> \textcolor{Red3}{\<ناهدة>} \<بس أنا اللي سمعته من>}\\
            {\small \<خطيبة عاصي يتيمة الأب و الأم>}
        \end{tabular} &
        \begin{tabular}{r}
            {\small \<بس أنا اللي سمعت من> \textcolor{Red3}{\<ناحية>} \<أنه>}\\
            {\small \<خطيبة عاصي، يتيمة الأب والأم>}
        \end{tabular} & M4T-v1\\

        Unr. Pron. & UAE &
        \begin{tabular}{r}
            {\small \<ما خليت هل البناية يرقدون، >} \\
            {\small \<تدقدق عليهم> \textcolor{Red3}{\<يالس>}}\\
        \end{tabular} &
        \begin{tabular}{r}
            {\small \textcolor{Red3}{\<يا صدق>} \<ما حليت هالبناية يرقدون>}\\
            {\small \<دق عليهم>}
        \end{tabular} & W-L-v3\\
        
        Alt. Orthog. & JOR &
        \begin{tabular}{r}
            {\small \<ليش، أنت رح> \textcolor{Red3}{\<بتقلي>} \<وبعدك>}\\
            {\small \<تجنني؟>}
        \end{tabular} &
        \begin{tabular}{r}
            {\small \<ليش؟ انت راح> \textcolor{Red3}{\<بتقول لي>} \<وبعدك>}\\
            {\small \<تجنني؟>}
        \end{tabular}
        & W-L-v3\\
        \bottomrule
    \end{tabular}
}
\caption{Examples for the different error categories observed during error analysis. Dia.: Dialect. Trans.: Translatoin. Unr.: Unrecognized. Pron.: Pronunciation. Alt.: Alternative.}
\label{tab:error-analysis-examples}
    
\end{table*}
\begin{table*}[]
    \centering
    \begin{tabular}{llccccc}
    \toprule
    Model & Error Type & Algeria & Jordan & Palestine & UAE & Yemen\\
    \midrule
    \multirow{7}{*}{SM4T-L-v2} & Total err. count & 86 & 24 & 20 & 106 & 124\\
    & Hallucination (\%) & 20.9 & 37.5 & 55.0 & 29.3 & 18.6\\
    & Deterioration (\%) & 26.7 & 33.3 & 20.0 & 26.4 & 28.2\\
    & Empty (\%) & 1.2 & 0.0 & 0.0 & 0.0 & 0.0\\
    & Incomplete (\%) & 8.1 & 4.2 & 5.0 & 1.89 & 0.8\\
    & MSA translation (\%) & 31.4 & 8.3 & 5.0 & 2.8 & 0.8\\
    & Dia. inaccuracies (\%) & 19.8 & 16.7 & 20.0 & 41.5 & 51.6\\
    \midrule
    \multirow{7}{*}{W-L-v3} & Total err. count & 87 & 27 & 16 & 104 & 111\\
    & Hallucination (\%) & 32.2 & 18.5 & 25.0 & 28.9 & 21.6\\
    & Deterioration (\%) & 31.0 & 33.3 & 43.8 & 36.5 & 34.2\\
    & Empty (\%) & 0.0 & 0.0 & 0.0 & 0.0 & 0.0\\
    & Incomplete (\%) & 1.2 & 7.4 & 0.0 & 5.8 & 6.3\\
    & MSA translation (\%) & 23.0 & 33.3 & 12.5 & 10.6 & 8.1\\
    & Dia. inaccuracies (\%) & 18.4 & 11.1 & 18.8 & 19.2 & 29.7\\
    \midrule
    \multirow{7}{*}{W-M} & Total err. count & 90 & 59 & 20 & 253 & 213\\
    & Hallucination (\%) & 35.6 & 28.8 & 40.0 & 35.2 & 33.3\\
    & Deterioration (\%) & 31.1 & 22.0 & 35.0 & 37.2 & 26.3\\
    & Empty (\%) & 0.0 & 0.0 & 0.0 & 0.0 & 0.0\\
    & Incomplete (\%) & 13.3 & 33.9 & 10.0 & 17.8 & 26.3\\
    & MSA translation (\%) & 14.4 & 15.3 & 10.0 & 6.7 & 4.2\\
    & Dia. inaccuracies (\%) & 6.7 & 5.1 & 5.0 & 6.3 & 13.2\\
    \midrule
    \multirow{7}{*}{HuBERT} & Total err. count & 28 & 4 & 9 & 75 & 61\\
    & Hallucination (\%) & 14.3 & 25.0 & 22.2 & 22.7 & 37.7\\
    & Deterioration (\%) & 57.1 & 50.0 & 55.6 & 68.0 & 60.7\\
    & Empty (\%) & 10.7 & 0.0 & 11.1 & 5.3 & 16.4\\
    & Incomplete (\%) & 3.6 & 0.0 & 0.0 & 1.4 & 4.9\\
    & MSA translation (\%) & 0.0 & 0.0 & 0.0 & 0.0 & 0.0\\
    & Dia. inaccuracies (\%) & 14.3 & 25.0 & 11.1 & 2.7 & 3.3\\
    \midrule
    \multirow{7}{*}{\makecell{DW-32-16\\(Ours)}} & Total err. count & \textbf{12} & \textbf{3} & \textbf{4} & \textbf{39} & \textbf{50}\\
    & Hallucination (\%) & 50.0 & 66.7 & 50.0 & 20.5 & 12.0\\
    & Deterioration (\%) & 25.0 & 0.0 & 25.0 & 23.1 & 20.0\\
    & Empty (\%) & 0.0 & 0.0 & 0.0 & 0.0 & 2.0\\
    & Incomplete (\%) & 8.3 & 0.0 & 25.0 & 2.6 & 4.00\\
    & MSA translation (\%) & 8.3 & 33.3 & 0.0 & 2.6 & 4.0\\
    & Dia. inaccuracies (\%) & 8.3 & 0.0 & 0.0 & 51.3 & 58.0\\
    \bottomrule
        \end{tabular}
    \caption{Error analysis statistics of different systems evaluated on our in-house data. Err.: Error. Dia.: Dialectal.}
    \label{tab:error-stats}
\end{table*}

\end{document}